\documentclass[10pt,conference,compsoc]{IEEEtran}
\ifCLASSOPTIONcompsoc
  \usepackage[nocompress]{cite}
\else
  \usepackage{cite}
\fi
\ifCLASSINFOpdf
  \usepackage[pdftex]{graphicx}
  \DeclareGraphicsExtensions{.pdf,.jpeg,.png}
\else
\fi
\usepackage{hyperref}
\hypersetup{
  hidelinks, 
}

\usepackage{gensymb}

\begin{document}
\urlstyle{same}
%
\title{Data-driven HR\\R\'esum\'e Analysis Based on Natural Language Processing and Machine Learning}
%
%
%
%
\author{\IEEEauthorblockN{Tim~Zimmermann}
\IEEEauthorblockA{Hasso~Plattner~Institute\\
Potsdam, Germany\\
tim.zimmermann@student.hpi.de}%
\and
\IEEEauthorblockN{Leo~Kotschenreuther}
\IEEEauthorblockA{SAP Labs\\
Palo Alto, USA\\
l.kotschenreuther@sap.com}%
\and
\IEEEauthorblockN{Karsten~Schmidt}
\IEEEauthorblockA{SAP Labs\\
Palo Alto, USA\\
karsten.schmidt01@sap.com}%
}

\IEEEtitleabstractindextext{%
\begin{abstract}
Recruiters usually spend less than a minute looking at each r\'esum\'e when deciding whether it's worth continuing the recruitment process with the candidate. Recruiters focus on keywords, and it's almost impossible to guarantee a fair process of candidate selection. The main scope of this paper is to tackle this issue by introducing a data-driven approach that shows how to process r\'esum\'es automatically and give recruiters more time to only examine promising candidates. Furthermore, we show how to leverage Machine Learning and Natural Language Processing in order to extract all required information from the r\'esum\'es. Once the information is extracted, a ranking score is calculated. The score describes how well the candidates fit based on their education, work experience and skills. Later this paper illustrates a prototype application that shows how this novel approach can increase the productivity of recruiters. The application enables them to filter and rank candidates based on predefined job descriptions. Guided by the ranking, recruiters can get deeper insights from candidate profiles and validate why and how the application ranked them. This application shows how to improve the hiring process by giving an unbiased hiring decision support.
\end{abstract}

}

\maketitle

\IEEEdisplaynontitleabstractindextext

%
\IEEEpeerreviewmaketitle

\section{Motivation}

Data-driven HR is the current trend in HR departments to replace the outdated notion of support function and turning HR into a pro-active counselor and education partner within a corporate environment. Especially, the `war for talent' and the amount of applications for open positions lead to new dimensions in processing candidate profiles and finding the best match \cite{xApplicationsForSinglePosition}. Recruiters and hiring managers can easily be biased or accidentally applying `filters'\footnote{E.g., only looking for a specific keyword or degree; and missing domain knowledge for a position to fill} on candidates without having a full 360\degree \,view on an individual candidate. Moreover, when having multiple profiles and multiple positions to fill, the problem of matching internal and external candidates is multiplying the necessary efforts. Candidates can easily flood recruiting inboxes via online channels. This is important, because according to Erica Breuer, tailored r\'esum\'es are still the most effective way to apply for a job \cite{differencesLinkedinresume}. This information flooding is maybe one of the reasons, why recruiters often ignore candidates that did not explicitly apply for a position instead of actively seeking them\footnote{Job hunting on online portals, such as linkedin is mostly focusing on professionals}. Having the right tools to objectively judge and rank candidates could help to a) find the best match and b) process more potential candidates.

In this research prototype, we leveraged current state-of-the-art technology in natural language processing (NLP) and machine learning (ML) to demonstrate how data-driven HR can significantly improve the quality and speed of the whole recruiting process.

\begin{figure}[!t]
\centering
\includegraphics[width=3.5in]{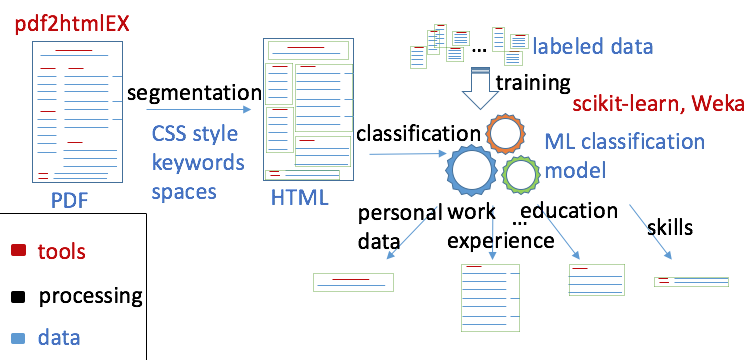}
\caption{From Document to Information: First we convert the PDF documents to HTML. Based on HTML structure, layout, and the content, we identify sections (personal, work experience, education and skills) using a pre-trained ML classifier.}
\label{fig:textExtraction}
\end{figure}

\section{From Document to Information}\label{section:documentToInformation}

\begin{figure*}[!t]
\centering
\includegraphics[width=7in]{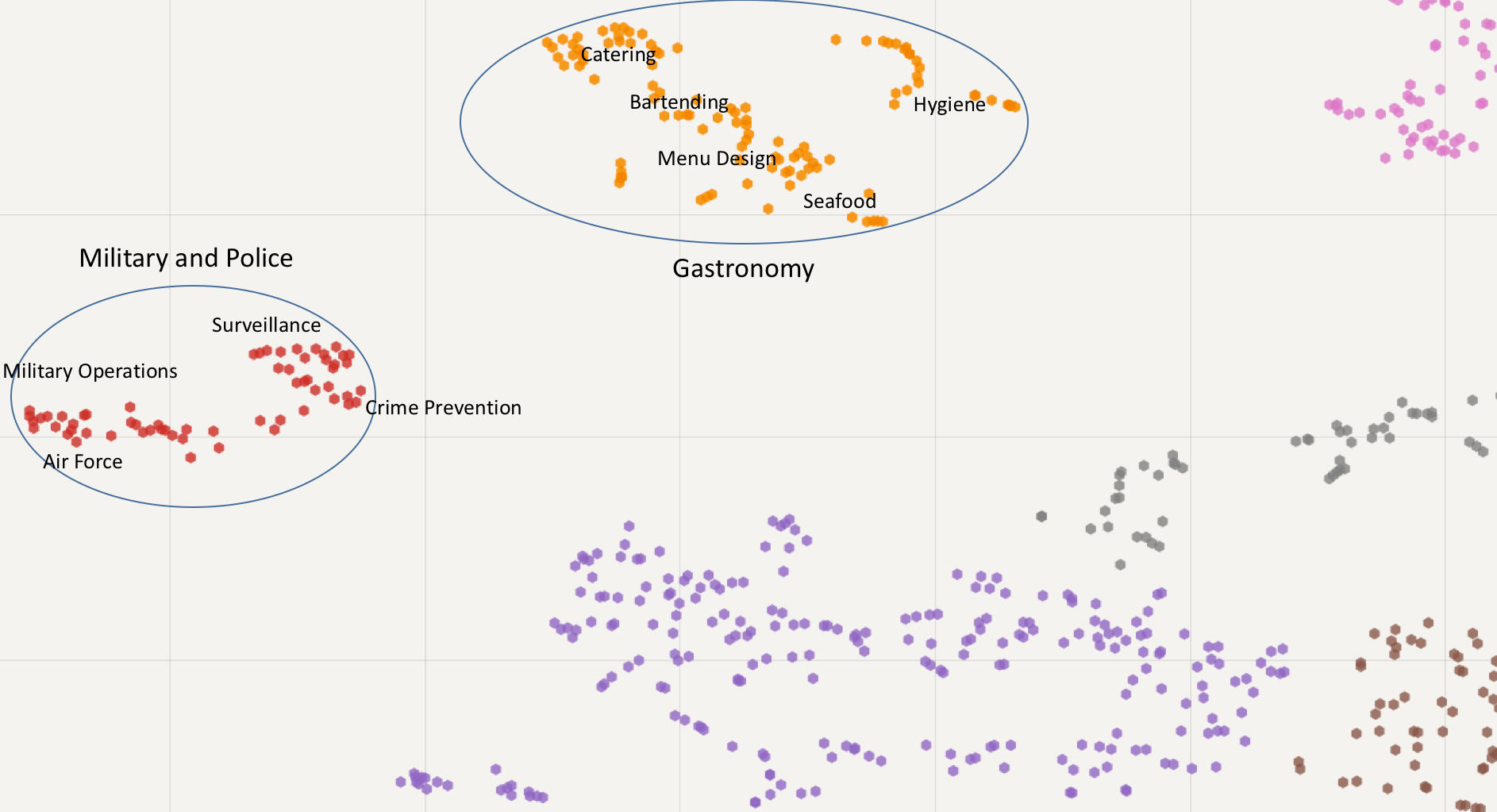}
\caption{This map shows clusters of related skills derived using word-embedding\cite{wordEmbedding}. Each dot represents a particular skill, the coloring represents different skill clusters.}
\label{fig:relatedSkillsClusters}
\end{figure*}

Most external\footnote{External - outside of an organization} job seekers have to provide multiple documents to prove their education, their work eligibility, language certificates, formal trainings, and a r\'esum\'e stating prior work experience, education, awards, skills and more. We observed that most r\'esum\'es are provided as a PDF document and leveraged their typical structured nature. Because, simple text extraction (e.g., xpdf's pdf2text\cite{xpdf}) may result in mixed segments (see Fig \ref{fig:textExtraction}), leveraging layout information is crucial. To correctly identify all entities in a r\'esum\'e, we employed segment-specific processors. For instance, once we identified the personal section or skill section, we instruct the processing pipeline accordingly. For the segmentation we identify headlines, extract font information, positioning, keywords, and spaces. Once the r\'esum\'e is split a pre-trained classifier\footnote{We used various SVM, RF, DT models and compared their performance} helps to correctly identify a section and route it to the right processor.

Each processor is using Named Entity Recognition (NER) to label locations, institutes, names, titles, and date information. Given the huge variety of date spellings, a separate regular expression step is necessary to normalize them. For instance, `Summer 2015', `Present/Now' or `2004,10 - 2005,9' are some of the non-standard instances which will be replaced with a normalized form `mm/dd/yyyy' or a duration type, respectively.

When inspecting segments like work experience, we re-apply the segmentation logic to split into individual career steps before labeling employer, dates, and role.

To ensure consistent quality, a combination of NLP/ML and well-defined rules helps to label education experience. The degree identification is based on a normalized bachelor, master, doctoral degree ranking with a certain spelling variation tolerance. Contrary, the segment identification is solely ML classification based.

Eventually, the skills section requires a customized processing, too. We pre-trained a skill co-occurrence model using word-embedding\cite{wordEmbedding} based on a large corpus of 800\,k profiles. For an easy validation and visualization we reduced the high dimensionality from hundred down to two using Stochastic Neighborhood Embedding (t-SNE\cite{tSNE}) and automatically identified clusters (see Fig \ref{fig:relatedSkillsClusters}) of related skills.

The whole pre-processing pipeline transforms a r\'esum\'e document into a set of structured information entities that can easily be processed.

\section{From Information to Knowledge}

The goal is to get a complete picture of a candidate's fit to a specific job position. Thus, we need to combine multiple dimensions of information, such as skills, education, and work experience. In this section we will show how we measure fit on each dimension and how we can merge them to gain a final ranking score representing a candidate's fit.

\subsection{Scoring}\label{subsection:scoring}

In our prototype, each candidate is assigned a score between 0 and 100 indicating its match to a given job description. The score is the weighted average of the three categories: education, work experience, and skills. By default, the skills score is twice as important as the other two.

\subsection{Education}

The education score is based on academic degree as well as the university's ranking. For university ranking we used: Times Higher Education \cite{TimesHigherEducation} and QS \cite{QS}. Both of them include a score between 0 and 100, with 100 being the best\footnote{For THE 100 means `perfect' and for QS 100 means `first'}. We use the average of both scores. If a university is not listed in one of these rankings, its score for that ranking is considered to be 0. We know that this may be unfair, but almost all universities we checked for our prototype were present in at least one of the rankings.
By default, the degree score is a constant number: 20 (Bachelor), 35 (Master) or 50 (Doctoral).
The sum of degree score and university ranking score leads to the final education score.
Note, that for scoring only the most recent university and degree are taken into account instead of previous ones. One of the reasons is that certain school types, e.g., High Schools, are not considered for university rankings which would result in 0 scoring. Eventually, candidates with multiple entries may have a couple of low (or zero) scores. An alternative would be to have a more fine-grained weighting based on duration/degree and even courses taken if available or to only consider the `top' scorer. Nevertheless, the goal of the prototype is to show how external data can objectively improve judgment about education qualities.

\subsection{Work Experience}

Our second score is for work experience that depends on duration of employment as well as on an employer score. Additionally, the more recent the employment the more this employment contributes to the work experience score. For simplification, each month of experience is worth one point.

It is extremely difficult to rank employers. Not only because the lack of data for each and every company out there (especially start-ups and SMEs), but also because there is not a single criteria to rank them. Moreover, how much does a ranking criteria such as `revenue' or `number of employees' support an individual's fit and level of experience. In order to have at least some employer `quality' metric in our ranking, we relied on employer's prior selectivity on hiring. We used the training set of 800\,k profiles introduced in Section \ref{section:documentToInformation} in order to evaluate the career progress of employees based on their current employer. Since we don't have enough data to take the complete career into account, we take the average education score of its employees as employer score. The final score is the sum of the experience points and the average of the weighted employer scores. It is limited to 100 points.

\subsection{Skills}

The skill score is the average scores of all desired skills. To calculate a specific skill score, we match the skill to the skill set of the candidate. For each candidate skill, the distance (see Section \ref{section:documentToInformation}) to the desired skill is calculated and the skill with the shortest distance is identified. The score for this desired skill is calculated as follows:
\begin{equation}
score = score_{match} - \alpha * distance
\end{equation}
Whereas $score_{match}$ is a constant, which reflects the score for exact matching skills of distance zero. Other skills are based on distance punishment regulated by the $\alpha$ parameter.

\section{Prototype Application}

To experience the potential of our application, we built a prototype that allows for filtering, ranking, and comparison of candidates. In \cite{video} we made a screencast available showing all major capabilities.

\subsection{Ranking Candidates}

Candidates are assigned multiple scores as described in Section \ref{subsection:scoring}. These scores are used to sort them based on ranking in various visualizations (see Section \ref{subsection:comparingCandidates}) of candidates. Either a weighted overall ranking can be applied or a fine-grained focused ranking based on individual skills, education, or work experience.

\begin{figure}[!t]
\centering
\includegraphics[width=2.5in]{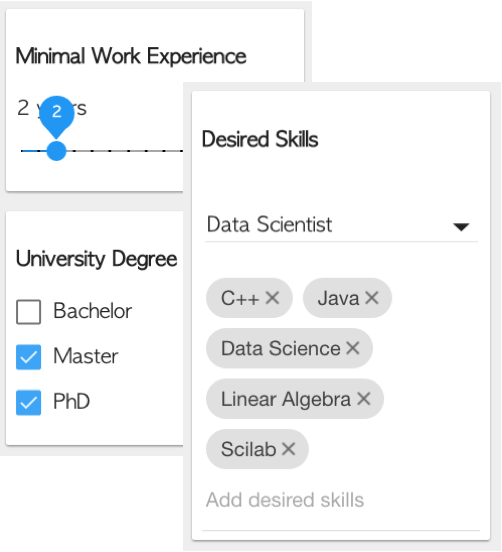}
\caption{Various ways for candidate filtering: minimal work experience in years, most current university degree and the mix of desired skills.}
\label{fig:filter}
\end{figure}

\begin{figure*}[!t]
\centering
\includegraphics[width=7in]{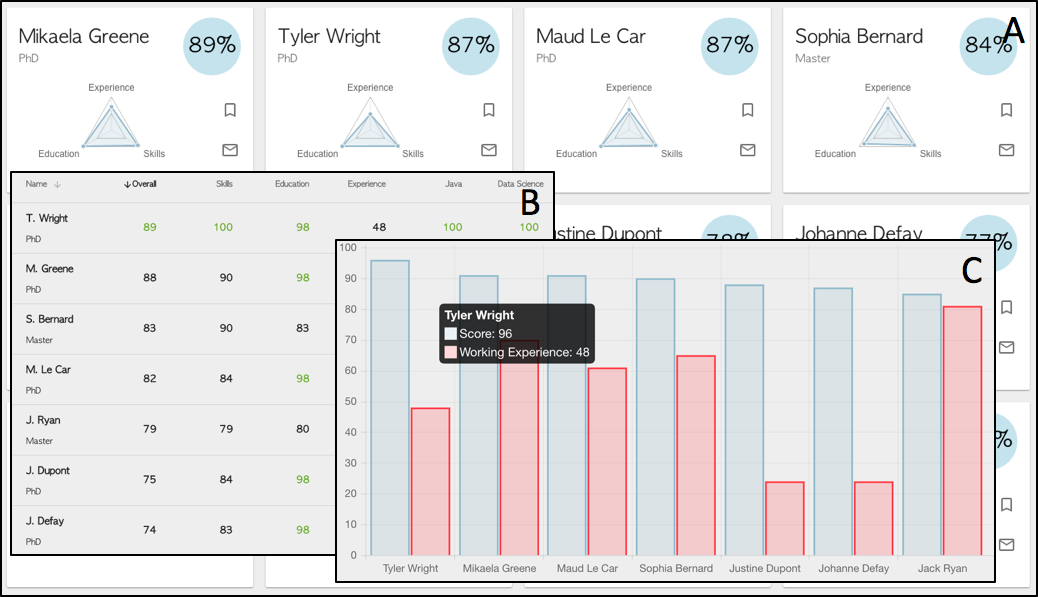}
\caption{Different views for ranking and comparing candidates: A shows the card view highlighting most important information (i.e., how well everyone fits the search requirements) for each candidate. B shows the score view which enables the user to inspect each specific score which plays a role in the overall score. Subfigure C shows the score chart. It displays the overall skill scores and the overall work experience scores ordered decreasingly by the skill scores. A screencast showcasing these features is available in \cite{video}.}
\label{fig:cardsScoresScoreChart}
\end{figure*}

\subsection{Filtering Candidates}

To filter candidates, the user\footnote{Recruiter or hiring manager} has different options (see Figure \ref{fig:filter}). One option is to filter by required degree(s). Another option is to specify the minimum number of years of work experience. Note, filtering does not affect the scoring, it just helps to reduce the pool size of candidates.

The user can also select any ensemble of skills to specify a desired candidate profile. Skill selection is supported by an auto-complete list, which is based on the large corpus of 800\,k profiles described in Section \ref{section:documentToInformation}. 

Note, skill scoring and ranking relies on the co-occurrence model on our training profiles. The benefit of that is that new skills automatically appear and mapped as long as new skill profiles are added. The downside is that we may experience a slight delay of new skills being correctly mapped, because it requires some occurrences of these new skills. 

Additionally, users can add related skills at any time and even start by using pre-defined job templates.

\subsection{Comparing Candidates}\label{subsection:comparingCandidates}
The tool gives the user multiple ways of displaying the candidate pool.

\subsubsection{Cards}

The default way of displaying the candidates is the card view shown in Figure \ref{fig:cardsScoresScoreChart}\,A. Every candidate is represented by a card containing the most important aspects, such as name, score, and most recent degree.

Each card also has a chart that shows the score broken down in the three categories (education, work experience, and skills). The user can quickly see which category a candidate excels in or falls behind. Generally, the bigger the area of the triangle in the chart, the better the candidate's score is relative to all other candidates.

Convenience features, like bookmarking of candidates or direct contact options are present as well.

\subsubsection{Scores}

A tabular view of scores is depicted in Figure \ref{fig:cardsScoresScoreChart}\,B. This view lets the user quickly evaluate how a candidate fits w.r.t. specific skills. There is a column for each of the three main categories as well as all the desired skills. To highlight the top scorers for each score, the top ten percentage is colored green. The user also has the option to sort by a specific category or skill.

\subsubsection{Score Chart}

The score chart (see Figure \ref{fig:cardsScoresScoreChart}\,C) displays the overall skill scores and overall work experience scores of the candidates. It is ordered decreasingly by the skill scores. If two candidates have the same skill score, the candidate with the better work experience score comes first. The advantage of this view is that the user can visually compare how candidates perform in the mentioned score parts in particular. For instance, the example shows that skill fit is fairly similar for all the candidates, but the score for work experience is quite different. This helps to navigate candidate filtering by visually guiding towards distinctive dimensions.

\begin{figure*}[!t]
\centering
\includegraphics[width=7in]{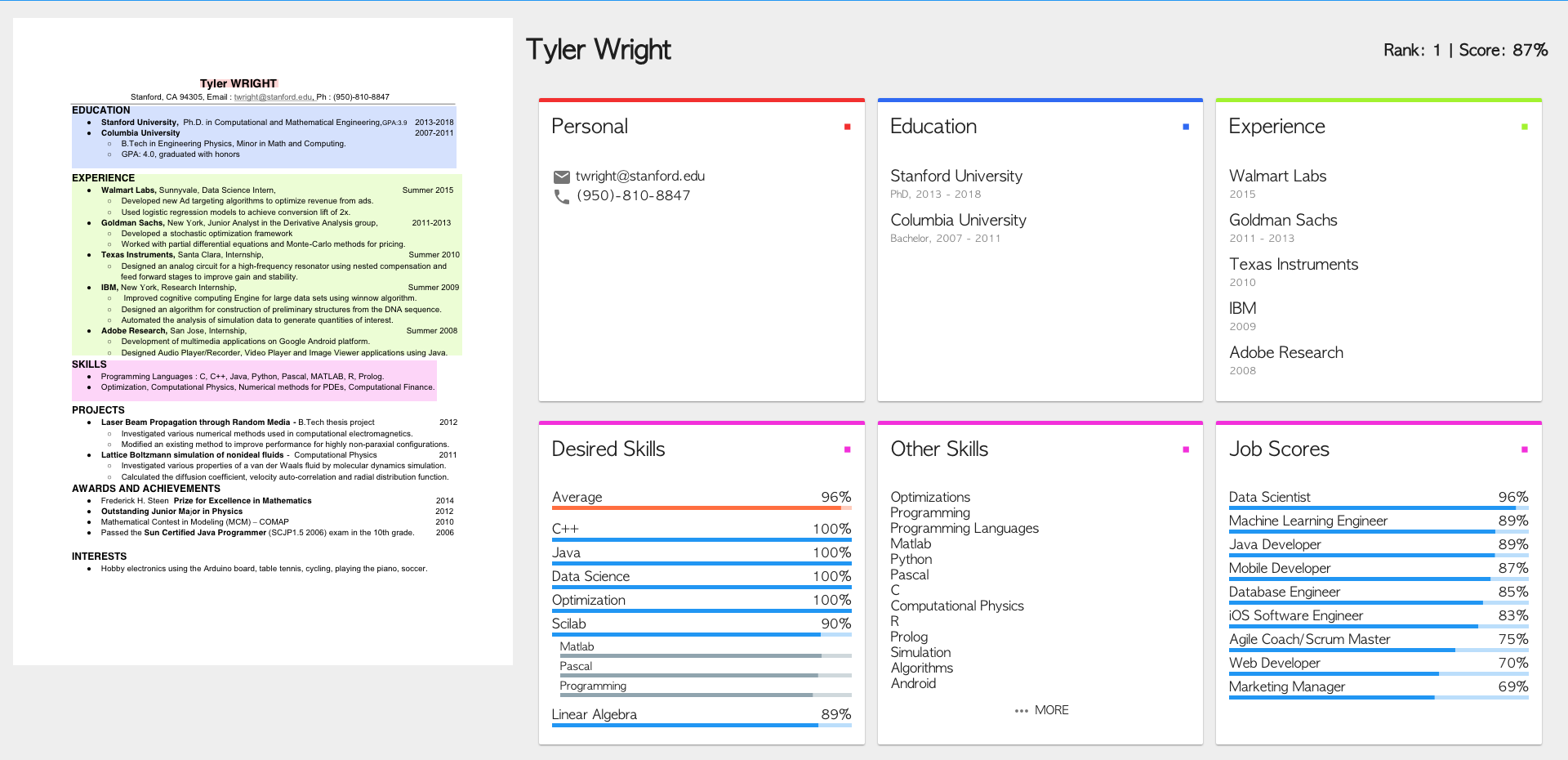}
\caption{Profile View of a candidate showing the original r\'esum\'e on the left and detail cards on the right. Any card content can be hovered to highlight relevant appearances in the document. Except for the Job Scores card, all cards are specific to the current r\'esum\'e and the previously defined job profile. }
\label{fig:profile}
\end{figure*}

\subsection{Inspecting Candidates}

One of the main features of the application is to analyze and explain the match between candidate skill set and job profile requirements. On the profile page (see Figure \ref{fig:profile}), the original r\'esum\'e is embedded on the left, while the extracted information is grouped in cards on the right.

\subsubsection{Context}

Hovering over an item of the extracted information cards, e.g., a skill, all occurrences of that skill are highlighted in the original r\'esum\'e. This provides the user with the ability to easily validate and learn more about the context of that skill. For instance, projects or classes this skill was applied in.

\subsubsection{Related Skills - Desired Skills Card}
Not only does the desired skills card visualize skill match quality for job profile skills, but each skill can also be expanded to show the top similar skills and the degree of similarity. For instance, Scilab is only a 90\,\% match. Since this skill is not explicitly mentioned in the r\'esum\'e, contributing skills, i.e., similar skills indicate knowledge in this area and based on our skill mapping technique a matching score below 100\,\% can be calculated.

\subsubsection{Match - Job Scores Card}

This card lists the top job profiles for this candidate based on our scoring fit measure. This is especially helpful for candidates who avoid applying for too many positions at the same time and recruiters who struggle to find the best candidate for a job. This card emphasizes if this candidate would be even a better fit for another job she has not applied for.

\section{Summary and next Steps}

This research prototype shows how to use a data-driven approach including multiple datasources in order to guide a user in matching candidates and jobs. Using ML and NLP, it is possible to build a pipeline that first extracts all the relevant information from r\'esum\'es and provides them in a structured way. Once r\'esum\'es are processed, external data for employers and educational institutes are included as well to calculate candidate matches. Recruiters can tailor their search and filtering to specific job roles. Additionally, there are several options and dimensions to compare candidates. Eventually, the application allows for a detailed analysis of the r\'esum\'e to validate ranking and matching recommendations.

However, because we developed a prototype, there are potential next steps and areas to further improve. First of all, weights to calculate scores are not optimized yet. More specifically, user research is necessary to evaluate various configurations. Secondly, in this work, we focused on extracting the most important information from r\'esum\'es. In fact, there are much more qualities to collect and to assess, e.g., awards, courses, job details, and languages. In addition, assessing career performance could be extremely beneficial. Other criteria to take into account could be costs of hiring candidates, relocation, or training to fill the gaps. When hiring people it is important to know whether they are going to fit into the team and can adapt to the corporate culture. Given all the data points that are available today, there are many more extensions of our prototype presented in this work.

\ifCLASSOPTIONcompsoc
  \section*{Acknowledgments}
\else
  \section*{Acknowledgment}
\fi

The authors would like to thank the team from SAP Innovation Center, Palo Alto: Frank Blechschmidt, Fredrick Chew, Pascal Crenzin, Stephan Haarmann, Michael Janke, Jaeyoon Jung, Roger Li, Bhumi Patel, Stefan Selent and Ren\'e Springer.

\ifCLASSOPTIONcaptionsoff
  \newpage
\fi

\end{document}